\documentclass[letterpaper]{article} 
\usepackage{aaai25}  
\usepackage{times}  
\usepackage{helvet}  
\usepackage{courier}  
\usepackage[hyphens]{url}  
\usepackage{graphicx} 
\usepackage{natbib}  
\usepackage{caption} 
\usepackage{algorithm}
\usepackage{algorithmic}

\usepackage{newfloat}
\usepackage{listings}
\DeclareCaptionStyle{ruled}{labelfont=normalfont,labelsep=colon,strut=off} 
\floatstyle{ruled}
\newfloat{listing}{tb}{lst}{}
\floatname{listing}{Listing}
\usepackage{graphicx}
\usepackage{booktabs}
\usepackage{todonotes}
\usepackage{amsmath}
\usepackage{amsfonts} 
\usepackage{svg}
\usepackage{algorithmic}
\usepackage{algorithm}
\usepackage{amsmath}
\usepackage{amssymb}
\usepackage{mathtools}
\usepackage{amsthm}
\ifdefined\preamble\else

\newcommand{\h}{\mathbf{h}}

\newcommand{\R}{\mathbb{R}}
\newcommand{\C}{\mathbb{C}}
\newcommand{\preamble}{}
\fi

\title{Real-Time Recurrent Reinforcement Learning}
\author{
    Julian Lemmel,
    Radu Grosu
}
\affiliations{
    Vienna University of Technology\\
    DatenVorsprung GmbH

    julian.lemmel@tuwien.ac.at\\
    radu.grosu@tuwien.ac.at
}

\begin{document}

\maketitle

\begin{abstract}
We introduce a biologically plausible RL framework for solving tasks in partially observable Markov decision processes (POMDPs). 
The proposed algorithm combines three integral parts: (1) A Meta-RL architecture, resembling the mammalian basal ganglia; (2) A biologically plausible reinforcement learning algorithm, exploiting temporal difference learning and eligibility traces to train the policy and the value-function; (3) An online automatic differentiation algorithm for computing the gradients with respect to parameters of a shared recurrent network backbone.
Our experimental results show that the method is capable of solving a diverse set of partially observable reinforcement learning tasks. The algorithm we call real-time recurrent reinforcement learning (RTRRL) serves as a model of learning in biological neural networks, mimicking reward pathways in the basal ganglia.
\end{abstract}

%
 \begin{links}
     \link{Code}{https://github.com/FranzKnut/RTRRL-AAAI25}
\end{links}

\section{Introduction}
Artificial neural networks were originally inspired by biological neurons, which are in general recurrently connected, and subject to synaptic plasticity. These long-term changes of synaptic efficacy are mediated by locally accumulated proteins, and a scalar-valued reward signal represented by neurotransmitter concentrations (e.g. dopamine)~\citep{wise2004}. The ubiquitous backpropagation through time algorithm~(BPTT)~\citep{BPTT}, which is used for training RNNs in practice, appears to be biologically implausible, due to distinct forward and backward phases and the need for weight transport~\citep{bartunov2018}.
With BPTT, RNN-based algorithms renounce their claim to biological interpretation. 
Biologically plausible methods for computing gradients in RNNs do however exist. One algorithm of particular interest is random feedback local online learning (RFLO)~\citep{murray2019}, an approximate version of real-time recurrent learning (RTRL)~\citep{williams1989}. Similarly, Linear Recurrent Units (LRUs)~\cite{zucchet2023} allow for efficient computation of RTRL updates. 

Gradient-based reinforcement-learning (RL) algorithms, such as temporal-difference (TD) methods, have been shown to be sample efficient, and come with formal convergence guarantees when using linear function approximation \citep{sutton2018}. However, linear functions are not able to infer hidden state variables that are required for solving POMDPs. RNNs, can compensate for the partial observability in POMDPs by aggregating information about the sequence of observations. Model-free deep reinforcement learning algorithms, leveraging recurrent neural network architectures (RNNs), serve as strong baselines for a wide range of partially-observable Markov decision processes (POMDPs) \citep{ni2022}. 
Contemporary RL algorithms further renounce biological plausibility due to the fact that updates are computed after collecting full trajectories, when future rewards are known. 

\textit{The question we asked in this paper was whether using a biologically plausible method for computing the gradients in RNNs, such as RFLO, in conjunction with a biologically plausible online RL method, such as TD($\lambda$), would be able to solve partially observable reinforcement learning tasks.} 

Taking advantage of previous work, we are able to answer the above question in a positive fashion. In summary, our proposed approach consists of three basic building blocks: 

\begin{figure}
  \centering
  \includegraphics[width=0.9\linewidth]{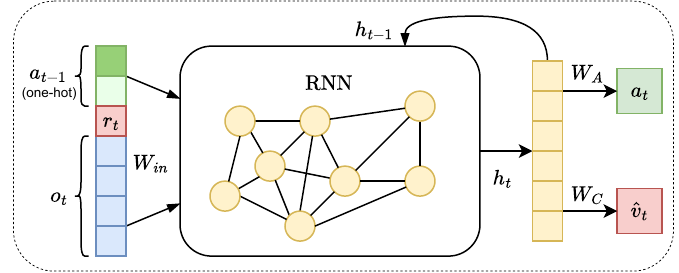}
 \caption{RTRRL uses a Meta-RL RNN-backbone which receives observation $o_t$, previous action $a_{t-1}$ and reward $r_{t}$, computing the latent vector $h_t$ from which the action $a_t$ and the value estimate $\hat v_t$ are computed via linear functions. 
 }
  \label{fig:shared}
\end{figure}

\begin{enumerate}
\item\textit{A Meta-RL RNN architecture}, resembling the mammalian basal ganglia depicted in figure \ref{fig:shared},
\item\textit{The TD($\lambda$) RL algorithm}, exploiting backwards-oriented eligibility traces to train the weights of the RNN.
\item\textit{Biologically-plausible RFLO or diagonal RTRL}, for computing the gradients of RNN-parameters online.
\end{enumerate}

\textit{We call our novel, biologically plausible RL approach real-time recurrent reinforcement learning (RTRRL).} 

In the Appendix, we argue in depth why BPTT is biologically implausible. Here, we summarize the two main objections: First, BPTT's reliance on shared weights between the forward and the backward synapses; Second, reciprocal error-transport, requiring propagating back the errors without interfering with the neural activity \citep{bartunov2018}; Third, the need for storing long sequences of the exact activation for each cell \citep{lillicrap2019a}. 

Previous work on Feedback Alignment (FA) \citep{lillicrap2016} demonstrated that weight transport is not strictly required for training deep neural networks. Particularly, they showed that randomly initialized feedback matrices, used for propagating back gradients to previous layers in place of the forward weights, suffice for training acceptable function approximators. They showed that the forward weights appear to align with the fixed backward weights during training.

Online training of RNNs was first described by \citet{williams1989}, who introduced real-time recurrent learning (RTRL) as an alternative to BPTT. More recently, a range of computationally more efficient variants were introduced \citep{tallec2018, roth2018, mujika2018, murray2019}. \citet{marschall2020} proposed a unifying framework for all these variants and assessed their biological plausibility while showing that many of them can perform on-par with BPTT. RFLO~\citep{murray2019}, stands out due to its biologically plausible update rule. LRUs~\cite{orvieto2023} have gained much attention lately, as they were shown to outperform Transformers on tasks that require long-range memory. Importantly, their diagonal connectivity reduces the complexity of RTRL trace updates, enabling them to compete with BPTT.~\citep{zucchet2023}.

Our second objection to biological plausibility of state-of-the-art RL algorithms is the use of multi-step returns in Monte-Carlo methods. Aggregating reward information over multiple steps helps reducing bias of the update target for value learning. However, in a biological agent, this requires knowledge of the future. 
While gathering some dust, TD($\lambda$) is fully biologically plausible due to the use of a temporal-difference target and backwards-oriented eligibility trace~\cite{sutton2018}.
Thus, the biological plausibility of RTRRL relies on three main building blocks: (1) The basal-ganglia inspired Meta-RL RNN architecture, (2) The pure backward-view implementation of TD($\lambda$), and (3) The RFLO automatic-differentiation algorithm or LRU RNNs trained with RTRL, as an alternative to BPTT.

This work demonstrates that online reinforcement learning with RNNs is possible with RTRRL, which fulfills all our premises for biologically plausible learning. We create a fully online learning neural controller that does not require multi-step unrolling, weight transport or communicating gradients over long distances. Our algorithm succeeds in learning a policy from one continuing stream of experience alone, meaning that no batched experience replay is needed. 
Our experimental results show that the same architecture, when trained using BPTT, achieves a similar accuracy, but entailing a worse time complexity. Finally, we show that the use of FA does not diminish performance in many cases.



%

\section{Real-Time Recurrent RL} 

%

In this section, we provide a gentle and self-contained introduction to each of the constituent parts of the RTRRL framework, namely the RNN models used, the online TD($\lambda$) actor-critic reinforcement learning algorithm, and RTRL as well as RFLO as a biologically plausible method for computing gradients in RNNs. We then put all pieces together and discuss the pseudocode of the overall RTRRL approach.


\paragraph{Continuous-Time RNN.} Introduced by \citet{CTRNN}, this type of RNN can be interpreted as rate-based model of biological neurons. In its condensed form, a CT-RNN with $N$ hidden units, $I$ inputs, and $O$ outputs has the following latent-state dynamics:
\begin{equation}
\label{eq:ctrnn}
h_{t+1} = h_{t}+\frac{1}{\tau}\left(-h_{t}+\varphi(W \xi_t)\right) \quad \xi_t=\begin{bmatrix}  x_{t} \\h_{t}\\ 1\end{bmatrix} \in \R^{Z}
\end{equation}
where $x_{t}$ is the input at time $t$, $\varphi$ is a non-linear activation function, $W$ a combined weight matrix $\in \R^{N\times X}$, $\tau$ the time-constant per neuron $\,{\in}\,\R^{N}$, and $\xi$ a vector $\in \R^{Z}$ with $Z\,{=}\,I\,{+}\,N\,{+}\,1$, the $1$ concatenated to $\xi_t$ accounting for the bias. The output $\hat y_{t} \in \R^O$ is given by a linear mapping $\hat y_{t}=W_{out}h_t$. The latent state follows the ODE defined by $\dot h_t = \tau^{-1}(-h_{t}+\varphi(W \xi_t))$, an expression that tightly resembles conductance-based models of the membrane potentials in biological neurons~\cite{gerstner2014}. 

\paragraph{Linear Recurrent Units (LRUs).} As a special case of State-Space Models \citep{gu2021}, the latent state of this simple RNN model is described by a linear system:
\begin{equation}
\label{eq:lru}
\dot h_t=A h_t+B x_t \qquad \quad y_t=\Re\left[C h_t\right]+D  x_t
\end{equation}
where $A$ is a diagonal matrix $\in \C^{N\times N}$ and $B, C, D$ are matrices $\in \C^{N\times I}, \C^{O\times N}$ and $\R^{O\times I}$ respectively. Note that the hidden state $h_t$ is a complex-valued vector $\in \C^N$ here. For computing the output $y_t$, the real part of the hidden state is added to the input $x_t$ at time $t$.
The name Linear Recurrent Unit was introduced in the seminal work of \citet{orvieto2023}. LRUs gained a lot of attention recently as they were shown to perform well in challenging tasks. The linear recurrence means that updates can be computed very efficiently. 




\paragraph{The Meta-RL RNN Architecture.} 
The actor-critic RNN architecture used by RTRRL is shown in figure~\ref{fig:shared}. It features a RNN with linear output layers for the actor and critic functions. At each step, the RNN computes an estimated latent state and the two linear layers compute the next action and the value estimate, from the latent state, respectively. Since the synaptic weights of the network are trained (slowly) to choose the actions with most value, and the network states are also updated during computation (fast) towards the same goal, this architecture is also called a Meta-RL. As shown by \citet{wang2018}, a Meta-RL RNN can be trained to solve a family of parameterized tasks where instances follow a hidden structure. They showed that the architecture is capable of inferring the underlying parameters of each task and subsequently solve unseen instances after training. Furthermore, they showed that the activations in trained RNNs mimic dopaminergic reward prediction errors (RPEs) measured in primates. RTRRL replaces the LSTMs used in \citet{wang2018} with CT-RNNs, allowing the use of RFLO as a biologically plausible method for computing the gradients of the network's parameters, or with LRUs which allow for efficient application of RTRL.
%
%
\paragraph{Temporal-Difference~Learning (TD).}   
TD Learning is a RL method that relies only on local information by bootstrapping \citep{sutton2018}. It is online, which makes it applicable to a wide range of problems, as it does not rely on completing an entire episode prior to computing the updates. After each action, the reward $r_t$, and past and current states $s_{t}$ and $s_{t+1}$ are used to compute the TD-error $\delta$:
\begin{equation}
\delta_t= r_t + \gamma \hat{v}_{\theta_t}(s_{t+1}) - \hat{v}_{\theta_t}(s_t)
\end{equation}
where $\hat v_\theta(s)$ are value estimates. The value-function is learned by regression towards the bootstrapped target. Accordingly, updates are computed by taking the gradient of the value-function and multiplying with the TD-error $\delta_t$: $\theta_{t+1} \gets \theta_t + \eta \delta_t \nabla_\theta \hat v_{\theta_t}(s_t)$, where $\eta$ is a small step size. 
\paragraph{Policy Gradient.} In order to also learn behavior, we use an actor-critic (AC) policy gradient method. In AC algorithms, the actor computes the actions, and the critic evaluates the actor's decisions. 
The actor (policy) is in this case a parameterized function $\pi_\varphi$ that maps state $s$ to a distribution of actions $p(a|s)$. The policy $\pi_\varphi$ is trained using gradient ascent, taking small steps in the direction of the gradient, with respect to the log action probability, multiplied with the TD-error. Particularly, the updates take the form $\varphi \gets \varphi + \alpha \delta \nabla_\varphi \log \pi_\varphi(a)$. Intuitively, this aims at increasing the probability for the chosen action whenever $\delta_t$ is positive, that is, when the reward was better than predicted. Conversely, when the reward was worse than expected, the action probability is lowered.
The TD-error is a measure for the accuracy of the reward-prediction, acting as RPE. Given its importance, it is used to update both the actor and the critic, acting as a reinforcement signal \citep{sutton2018}. Note the difference between the \emph{reward-signal} $r$ and the \emph{reinforcement-signal} $\delta$: if the reward $r$ is predicted perfectly by $\hat{v}_\theta$, no reinforcement $\delta$ takes place, whereas the absence of a predicted reward $r$ leads to a negative reinforcement $\delta$. 
Although AC was devised through mathematical considerations, the algorithm resembles the dopaminergic feedback pathways found in the mammalian brain.
\begin{figure}[t]
  \centering
  \includegraphics[width=\linewidth]{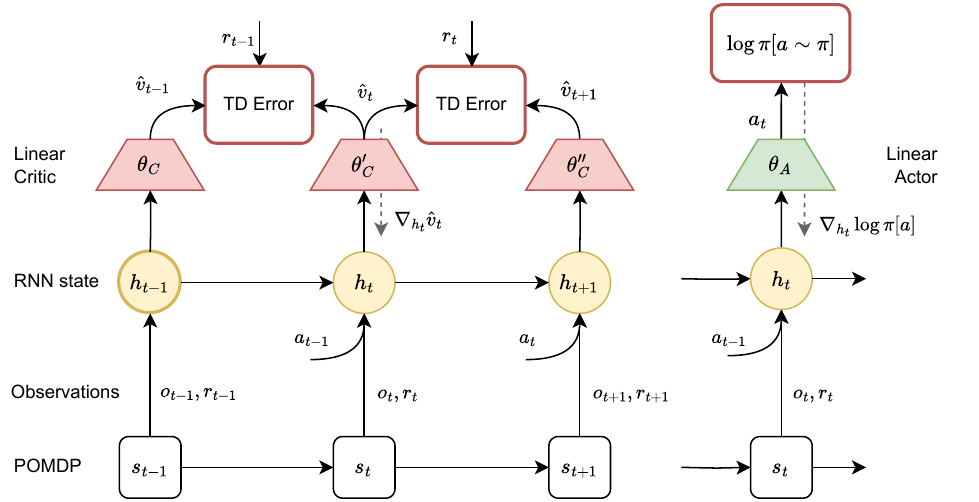}
  \caption{Schematics showing how gradients are passed back to the RNN (yellow). Gradients of the actor (red) and critic (green) losses are propagated back towards $h_{t}$ and $h_{t-1}$ respectively.}
  \label{fig:online_a}
\end{figure}
\paragraph{Eligibility Traces (ET).} The algorithm just described is known as TD(0). It is impractical when dealing with delayed rewards in an online setting, since value estimates need to be propagated backwards in order to account for temporal dependencies. 
%
ETs are a way of factoring in future rewards. The idea is to keep a running average of each parameter's contribution to the network output. This can be thought of as a short-term memory, paralleling the long-term one represented by the parameters. ETs unify and generalize Monte-Carlo and TD methods \citep{sutton2018}.
Particularly, TD($\lambda$) makes use of ETs. Weight updates are computed by multiplying the TD-error $\delta$ with the trace that accumulates the gradients of the state-value-function. The trace $e^\theta$ decays with factor $\gamma\lambda$ where $\gamma$ is the discount factor:
\begin{equation}
  e_{t}^\theta = \gamma\lambda\ e_{t-1}^\theta + \nabla_\theta \hat{v}_{\theta_t}(s_t) \qquad\qquad
  {\theta} \leftarrow {\theta} + \eta^\theta\ \delta_t\ e_t^\theta
\end{equation}
Since in RTRRL we use a linear value-function $\hat v_\theta (s_{t})\,{=}\, w^\top s_{t}$ with parameters $w$, like in the original TD($\lambda$), the gradient of the loss with respect to $w$ is simply $\nabla_{w}\hat v_\theta\,{=}\,s_{t}$. 
%
%
%
Linear TD($\lambda$) comes with provable convergence guarantees \citep{sutton2018}. However, the simplicity of the function approximator fails to accurately fit non-linear functions needed for solving harder tasks. Replacing the linear functions with neural networks will introduce inaccuracies in the optimization. However in practice, multi-layer perceptrons (MLPs) can lead to satisfactory results, for example on the Atari benchmarks \citep{daley2019}. 
The gradients of the synaptic weights in the shared RNN are computed in an efficient, biologically plausible, online fashion as discussed in the Introduction. To this end, RTRRL is using LRUs trained with RTRL, or CT-RNNs trained with RFLO, an approximation of RTRL. 
The gradients of the actor and the critic with respect to the RNN's hidden state are combined and propagated back using Feedback Alignment. In figure~\ref{fig:online_a}, we show how the gradients are passed back to the RNN for RFLO. 

%
%
%
%
\paragraph{Real-Time Recurrent Learning (RTRL).} RTRL was proposed by \citet{williams1989} as an RNN online optimization algorithm for infinite horizons. 
The idea is to estimate the gradient of network parameters during the feedforward computation, and using an approximate of the error-vector to update the weights in each step. Bias introduced due to computing the gradient with respect to the dynamically changing parameters is mitigated by using a sufficiently small learning rate $\eta$. The update rule used in RTRL is introduced shortly. 

Given a dataset consisting of the multivariate time-series $x_t\,{\in}\,\R^I$ of inputs and $y_t\,{\in}\,\R^O$ of labels, we want to minimize some loss function  $\mathcal L_{\theta}=\sum_{t=0}^{T} L_{\theta}(x_t,y_{t})$ by gradient descent. This is achieved by taking small steps in the direction of the negative gradient of the total loss:
\begin{equation}
\begin{aligned}
\Delta \theta =-\eta \nabla_{\theta}\mathcal L_\theta =-\eta\sum\nolimits_{t=0}^{T} \nabla_\theta L_{\theta}(t)
\end{aligned}
\end{equation}
We can compute the gradient of the loss as $
\Delta \theta(t)=\nabla_{\theta}L_{\theta}(t) = \nabla_{\theta}\hat y_{t}\nabla_{\hat y_{t}}L_\theta(t)$ with $\hat y_{t}$ being the output of the RNN at timestep $t$. When employing an RNN with linear output mapping, the gradient of the model output can be further expanded into $\nabla_{\theta_R}\hat y_{t} = \nabla_{\theta_{R}}h_{t} \nabla_{h_z} \hat y_t$. 
The gradient of the RNN's state with respect to the parameters $\theta$ is computed recursively. To this end, we define the immediate Jacobian $\bar J_{t}\,{=}\,\nabla_{\theta} f(x_{t},h_{t})$, with $\nabla_{\theta}$ being the partial derivative with respect to $\theta$. Likewise, we introduce the approximate Jacobian trace $\hat J_t \approx \frac{d}{d \theta} f(x_{t},h_{t})$, where $\frac{d}{d \theta}$ is the total derivative and $f(x_t,h_t)=\dot h_t$ for the respective RNN model used.
\begin{equation} 
\label{eq:RTRL}
\begin{aligned}
\hat J_{t+1} := \frac{d}{d\theta} h_{t+1} &= \frac{d}{d\theta}\left(h_{t} + f(x_{t}, h_{t}) \right) \\
 &= \hat J_{t}\left( \mathbb I + \nabla_{h_{t}} f(x_{t},h_{t})\right) + \bar J_t
\end{aligned}
\end{equation}
Equation \ref{eq:RTRL} defines the Jacobian trace recursively, in terms of the immediate Jacobian and a linear combination of the past trace $\hat J_{t}\nabla_{h_{t}} f(x_{t},h_{t})$, allowing to calculate it in parallel to the forward computation of the RNN. When taking an optimization step we can calculate the final gradients as: 
\begin{equation}
\label{eq:rnn_update}
\Delta \theta(t) = \hat J_t  \nabla_{h_{t}}L_{\theta(t)}=\hat J_t W_{out}^{\top}\varepsilon_{t}
\end{equation}
Note that RTRL does not require separate phases for forward and backward computation and is therefore biologically plausible in the time domain. 
However, backpropagating the error signal $W_{out}^{\top}\varepsilon_{t}$ still assumes weight sharing, and the communicated gradients $\hat J_{t}\nabla_{h_{t}} f(x_{t},h_{t})$ violate locality.
 
One big advantage of RTRL-based algorithms, is that the computation time of an update-step is constant in the number of task steps.
However, RTRL has complexity $\mathcal O(n^4)$ in the number of neurons $n$ compared to $\mathcal O(nT)$ for BPTT, where $T$ is the time horizon of the task. RTRL is therefore not used in practice. 
Note however, that due to the diagonal connectivity of LRUs, each parameter of their recurrent weight matrix influences just a single neuron. This means that the complexity of the RTRL update is reduced significantly, as shown by \citet{zucchet2023}. Furthermore, they demonstrated how LRUs trained with RTRL can be generalized to the multi-layer network setting. LRU-RNNs are therefore a viable option for efficient online learning.

\paragraph{Random Feedback Local Online Learning (RFLO).}
\label{sec:RFLO}
A biologically plausible variant of RTRL is RFLO \citep{murray2019}. This algorithm leverages the Neural ODE of CT-RNNs in order to simplify the RTRL update substantially, conveniently dropping all parts that are biologically implausible. RFLO improves biological plausibility of RTRL in two ways: 1. Weight transport during error backpropagation is avoided by using FA, 2. Locality of gradient information is ensured by dropping all non-local terms from the gradient computation.
Particularly, (\ref{eq:rnn_update}) is leveraged in order to simplify the RTRL update. For brevity, we only show the resulting update rule. A derivation can be found in the Appendix. 
\begin{equation} 
\label{eq:RFLO}
\hat J_{t+1}^W \approx (1 -\tau^{-1})\ \hat J _t^W  + \tau^{-1} \varphi'(W \xi_t)^\top \xi_t
\end{equation}
Weight transport is avoided by using FA for propagating gradients. Parameter updates are computed as $\Delta W(t) =\hat J_t^W B \varepsilon_{t}$ using a fixed random matrix $B$. Effective learning is still achievable with this simplified version as shown in \cite{murray2019, marschall2020}.
RFLO has time complexity $\mathcal O(n^2)$ and is therefore less expensive than BPTT, when the horizon of the task is larger than the number of neurons. The reader is referred to \citet{murray2019} and \citet{marschall2020} for a detailed comparison between RTRL and RFLO as well as other approximations.
%
\begin{algorithm}[ht]
    \caption{RTRRL}
    \label{alg:RTRL}
    \begin{algorithmic}[1]
	    \REQUIRE Linear policy: $\pi_{\theta_{A}}(a|h)$
	    \REQUIRE Linear value-function: $\hat v_{\theta_{C}}(h)$
	    \REQUIRE Recurrent layer: RNN$_{\theta_{R}}([o, a, r], h, \hat J)$
	    \STATE $\theta_{A}, \theta_{C}, \theta_{R} \gets $ initialize parameters
	    \STATE $B_{A}, B_{C} \gets $ initialize feedback matrices
		\STATE $h, e_{A}, e_{C}, e_R \gets \mathbf 0$ 
		\STATE $o \gets $ reset Environment
	    \STATE $h, \hat J \gets $RNN$_{\theta_{R}}([o, \mathbf 0, 0], h, \mathbf 0)$ 
	    \STATE $v \gets \hat v_{\theta_{C}}(h)$
		\WHILE{not done}
			\STATE $\pi \gets \pi_{\theta_{A}}(h)$
		    \STATE $a \gets $ sample($\pi$)
		    \STATE $o, r \gets$ take action $a$
			\STATE $h', \hat J' \gets $RNN$_{\theta_{R}}([o, a, r], h, \hat J)$ 
            \STATE $e_{C} \gets \gamma \lambda_{C} e_{C} + \nabla_{\theta_{C}} \hat v$
			\STATE $e_{A} \gets \gamma \lambda_{A} e_{A} + \nabla_{\theta_{A}} \log \pi[a]$
            \STATE $g_{C} \gets B_{C} \mathbf 1$
            \STATE $g_{A} \gets B_{A} \nabla_{\pi} \log\ \pi[a] $
            \STATE $e_R \gets \gamma \lambda_{R} e_{R} + \hat J  (g_{C} + \eta_A g_{A})$  
            
            \STATE $v' \gets \hat v_{\theta_{C}}(h')$ 
            \STATE $\delta \gets r + \gamma v' - v$ 
			
			\STATE $\theta_{C} \gets \theta_{C} + \alpha_C \delta e_{C}$   
			\STATE $\theta_{A} \gets \theta_{A} + \alpha_{A} \delta e_{A}$
			\STATE $\theta_{R} \gets \theta_{R} + \alpha_{R}  \delta e_R$
			\STATE $v \gets v', \quad h \gets h', \quad \hat J \gets \hat J'$
		\ENDWHILE
    \end{algorithmic}
\end{algorithm}
%
\subsection{Putting All Pieces Together}

\begin{figure*}[ht!]
  \centering
\includegraphics[width=.3\linewidth]{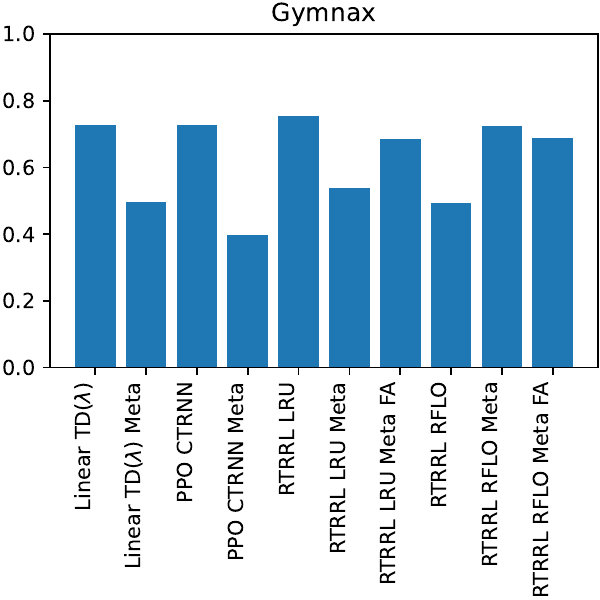}
\hspace{4em}
\includegraphics[width=0.3\linewidth]{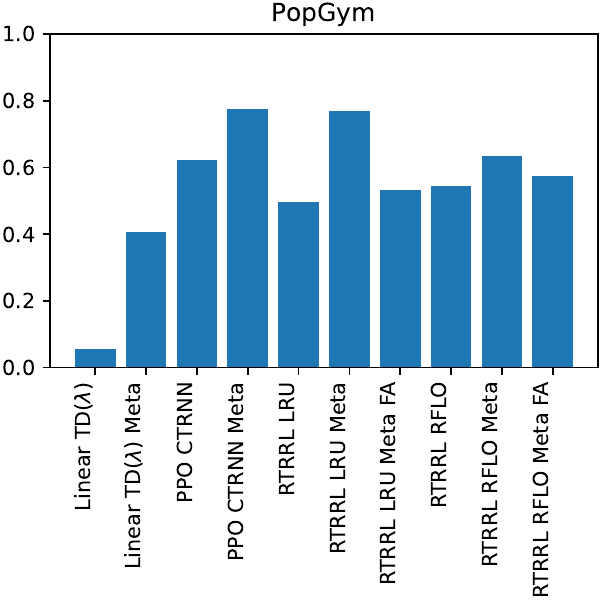}
  \caption{Boxplot of the combined normalized validation rewards achieved for 5 runs each on a range of different environments from the \texttt{gymnax}, and \texttt{popgym} packages. Depicted are results for RTRRL with CTRNNs and LRUs, each with and without FA. RTRRL-LRU-Meta and PPO-CTRNN-Meta perform best overall. Using FA always leads to diminished performance. Fully biologically plausible RTRRL-RFLO with FA often achieves on-par results.
  }
  \label{fig:boxplot}
  
\end{figure*}

Having described all the necessary pieces, we now discuss how RTRRL puts them together.
Algorithm~\ref{alg:RTRL} shows the outline of the RTRRL approach. Importantly, on line 11, the next latent vector $h'$ is computed by the single-layer RNN parameterized by $\theta_R$. We use the Meta-RL structure depicted in figure \ref{fig:shared}. Therefore, previous action, reward, and observation are concatenated to serve as input $[o,a,r]$. The approximate Jacobian $\hat J_t$ is computed as second output of the RNN step, and is later combined with the TD-error $\delta$ and eligibility trace $e_R$ of TD($\lambda$) to update the RNN weights. $\pi$ and $\hat v$ are the Actor and Critic functions parameterized by $\theta_A$ and $\theta_C$. 
We train the Actor and Critic using TD($\lambda$) and take small steps in direction of the log of the action probability $\pi[a]=\mathbf P\mathcal{N}(\theta_A^\top h')$ and value estimate $\hat v=\theta_V^\top h'$ respectively. The gradients for each function are accumulated using eligibility traces $e_{A,C}$ with $\lambda$ decay. Additionally, the gradients are passed back to the RNN through random feedback matrices $B_A$ and $B_C$ respectively. The eligibility trace $e_{R}$ for the RNN summarizes the combined gradient. The use of randomly initialized fixed backwards matrices $B_{A,C}$ in RFLO makes RTRRL biologically plausible.

The Jacobian in Algorithm~\ref{alg:RTRL} can also be computed  using RTRL. This reduces the variance in the gradients at a much higher computational cost. Similarly, we can choose to propagate back to the RNN by using the forward weights as in backpropagation. However, we found that the biologically plausible Feedback Alignment works in many cases.

The fully connected shared RNN layer acts as a filter in classical control theory, and estimates the latent state of the POMDP. Linear Actor and Critic functions then take the latent state as input and compute action and value respectively. We use CT-RNNs or LRUs as introduced in the previous section for the RNN body, where we compute the Jacobian $\hat J_t$ by using RFLO or RTRL as explained above. Extending RFLO, we derive an update rule for the time-constant parameter $\tau$. The full derivation can be found in the Appendix:
\begin{equation} 
\label{eq:RFLO_tau}
\begin{aligned}
\hat J_{t+1}^\tau \approx \hat J_t^\tau (1-\tau^{-1}) + \tau^{-2} \left(h_{t} - \varphi(W \xi_t)\right)
\end{aligned}
\end{equation}
%

The hyperparameters of RTRRL are $\gamma$,  $\lambda_{[A,C,R]}$, $\alpha_{[A,C,R]}$. Our approach does not introduce any new ones over TD($\lambda$) other than lambda and learning rate for the RNN. For improved exploration, we also compute the gradient of the action distribution's entropy, scaled by a factor $\eta_H$, and add it to the gradients of policy and RNN. In order to stay as concise as possible, this is omitted from algorithm \ref{alg:RTRL}. Further implementation details can be found in the Appendix.

\begin{figure*}[ht]
  \centering
  \includegraphics[width=0.66\linewidth]{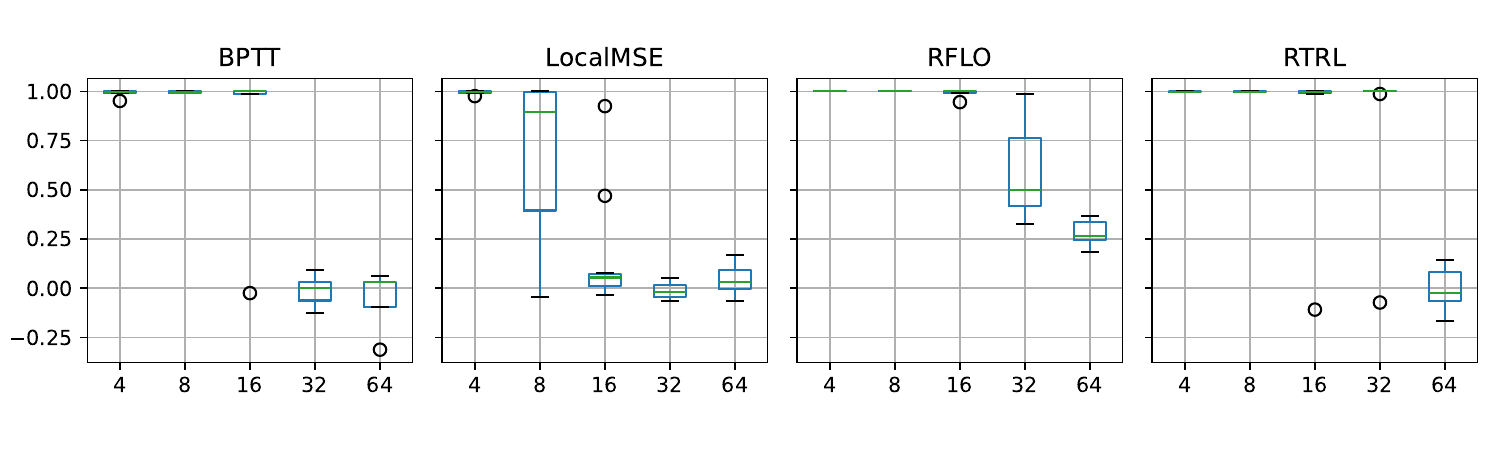}
  \hspace*{3em}
\includegraphics[width=0.2\linewidth]{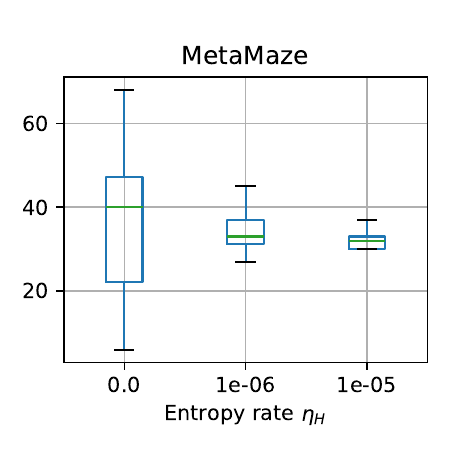}
  \caption{Left: Boxplots of 10 runs on \texttt{MemoryChain} per type of plasticity for increasing memory lengths. BPTT refers to PPO with LSTM, RFLO and RTRL denote the variants of RTRRL and LocalMSE is a naive approximation to RTRL. Right: Tuning the entropy rate is a trade-off of best possible reward vs. consistency as shown for the \texttt{MetaMaze} environment.}
  \label{fig:memory_length}
\end{figure*}

\begin{figure*}
  \centering
\includegraphics[width=\linewidth]{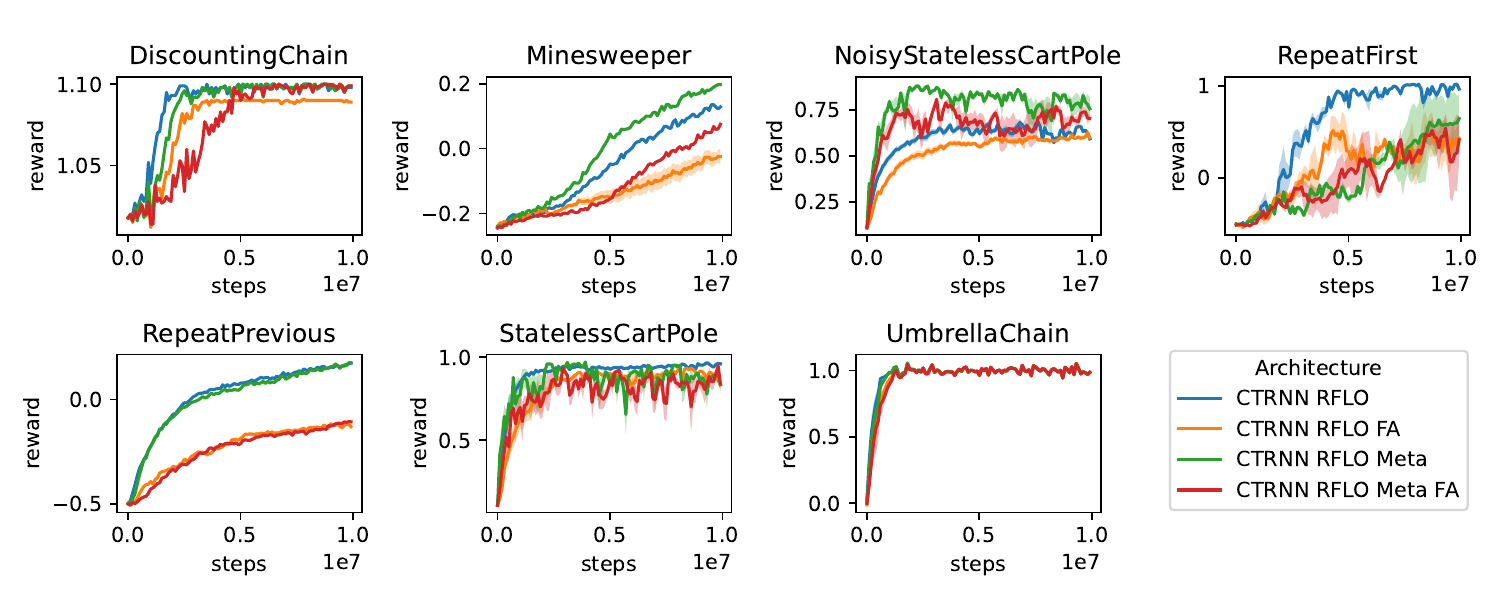}
  \caption{Left: Shown are the mean rewards aggregated over 5 runs each; shaded regions are the variance. While in many cases it does not make a difference, not using the Meta-RL architecture hampers performance in some cases. Using biologically plausible Feedback Alignment can lead to worse results, but more often than not it does not have a significant impact.
  }
  \label{fig:ablation}
  
\end{figure*}

\section{Experiments}
We evaluate the feasibility of our RTRRL approach by testing on RL benchmarks provided by the \texttt{gymnax}~\citep{gymnax2022github}, \texttt{popgym} ~\citep{morad2022} and \texttt{brax}~\citep{freeman2021} packages. The tasks comprise fully and partially observable MDPs, with discrete and continuous actions. 
As baselines we consider TD($\lambda$) with Linear Function Approximation, and Proximal Policy Optimization (PPO)~\citep{schulman2017} using the same RNN models but trained with BPTT. Our implementation of PPO is based on \texttt{purejaxrl} \citep{lu2022discovered}. We used a truncation horizon of 32 for BPTT. For each environment, we trained a network with 32 neurons for either a maximum of 50 million steps or until 20 subsequent epochs showed no improvement. The same set of hyperparameters, given in the Appendix, was used for all the RTRRL experiments if not stated otherwise. Importantly, a batch size of $1$ was used to ensure biological plausibility. All $\lambda$'s and $\gamma$ were kept at $0.99$, $\eta_H$ was set to $10^{-5}$, and the \texttt{adam}~\citep{kingma2015} optimizer with a learning rate of $10^{-3}$ was used.

For discrete actions, the outputs of the actor are the log probabilities for each action, and past actions fed to the RNN are represented in one-hot encoding. 
To obtain a stochastic policy for continuous actions, a common trick is to use a Gaussian distribution parameterized by the model output. We can then easily compute the gradient of $\log\pi[a]$. 

Figure \ref{fig:boxplot} shows a selection of our experimental results as boxplots. Depicted are the best validation rewards of 5 runs each. RTRRL-LRU achieves the best median reward in almost all cases, outperforming PPO. Interestingly, Linear TD($\lambda$) was able to perform very well on some environments that did require delayed credit-assignment, such as \texttt{UmbrellaChain}. At the same time it performed especially poor on POMDP environments such as \texttt{StatelessCartPole}. Finally, fully biologically plausible RTRRL-RFLO often shows comparable performance to the other methods and too can outperform PPO in some cases. One notable exception is \texttt{NoisyStatelessCartPole}, where RTRRL-RFLO performed best. This could hint at RFLO being advantageous in noisy environments. Investigating this theory however was outside of the scope of this paper. A second exception is \texttt{RepeatPrevious}, where RTRRL-RFLO performed surprisingly poorly for unknown reasons. Finally, we include results for both RTRRL versions when used with FA. 
\paragraph{Memory Length.}
We compared the memory capacity of RTRRL-CTRNN by learning to remember the state of a bit for an extended number of steps. The \texttt{MemoryChain} environment can be thought of as a T-maze, known from behavioral experiments with mice.~\citep{osband2020} 
The experiment tests if the model can remember the state of a bit for a fixed number of time steps. Increasing the number of steps increases the difficulty. We conducted \texttt{MemoryChain} experiments with 32 neurons, for exponentially increasing task lengths. A boxplot of the results is shown in figure \ref{fig:memory_length}. With this experiments we wanted to answer the question of how RTRL, RFLO, and BPTT compare to each-other. The results show that using approximate gradients hampers somewhat memory capacity. Quite surprisingly, RTRRL outperformed networks of the same size that were trained with BPTT. 

\begin{figure*}[ht]
  \centering
  \includegraphics[width=.9\linewidth]{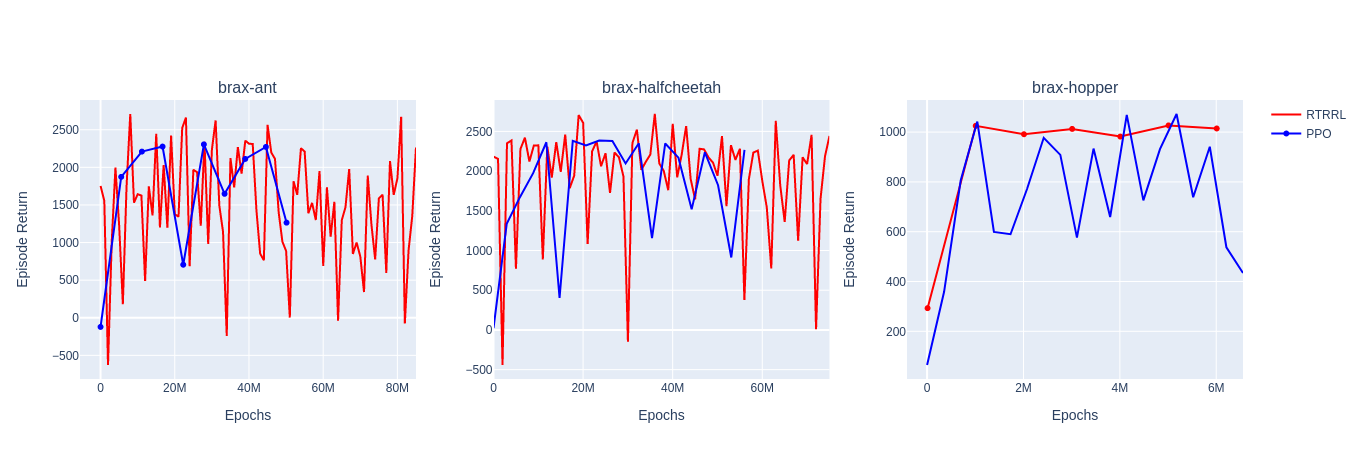}
  \caption{Rewards over training epochs for environments from the \texttt{brax} package, masked to make them POMDP. Shown are the best run for RTRRL vs. a tuned PPO baseline. RTRRL performance shows increased variance due to the batch size of 1.}
  \label{fig:brax} 
\end{figure*}
\paragraph{Ablation Experiments.}
In order to investigate the importance of each integral part of RTRRL, we conducted three different ablation studies with CT-RNNs. Note that for the first two ablation experiments, $\eta_H$ was kept at $0$.

\begin{itemize}
    \item Feedback Alignment: We conducted experiments to test whether using the biologically plausible FA would result in a decreased reward. While true for most environments, surprisingly, FA often performs just as well as regular backpropagation, as can be seen in figures \ref{fig:boxplot} and \ref{fig:ablation}.
    \item Meta-RL: We tried out RTRRL without the Meta-RL architecture and found that some tasks, such as \texttt{RepeatPrevious}, gain a lot from using the Meta-RL approach, as depicted in figure \ref{fig:ablation}. However, the results show that for many environments it does not matter.
    \item Entropy rate: One tunable hyperparameter of RTRRL is the weight given to the entropy loss $\eta_H$. We conducted experiments with varying magnitudes and found, that the entropy rate represents a trade-off between consistency and best possible reward. In other words, the entropy rate seems to adjust the variance of the resulting reward, as depicted in figure \ref{fig:memory_length} (right).
\end{itemize}

\paragraph{Physics Simulations.}
We masked the continuous action \texttt{brax} environments, keeping even entries of the observation and discarding odd ones, to create a POMDP.
For each environment, we ran hyperparameter tuning for at least 10 hours and picked the best performing run. Figure \ref{fig:brax} shows the evaluation rewards for RTRRL, compared to the tuned PPO baselines that were provided by the package authors. 

\section{Related Work}
\label{sec:related}

\citet{johard2014} used a cascade-correlation algorithm with two eligibility traces, that starts from a small neural network and grows it sequentially. This interesting approach, featuring a connectionist actor-critic architecture, was claimed to be biologically plausible, although they only considered feed-forward networks and the experimenta evaluation was limited to the \texttt{CartPole} environment.

In his Master's Thesis, \citet{chung2019} introduced a network architecture similar to RTRRL that consists of a recurrent backbone and linear TD heads. Convergence for the RL algorithm is proven assuming the learning rate of the RNN is magnitudes below those of the heads. Albeit using a network structure similar to RTRRL, gradients were nonetheless computed using biologically implausible BPTT.

\citet{ororbia2022} introduced a biologically plausible model-based RL framework called active predicitve coding that enables online learning of tasks with extremely sparse feedback. Their algorithm combines predictive coding and active inference, two concepts grounded in neuroscience. Network parameters were trained with Hebbian updates combined with component-wise surrogate losses. 

One approach to reduce the complexity of RTRL was proposed by \citet{javed2023}. Similar to \citet{johard2014}, a RNN is trained constructively, one neuron at a time, subsequently reducing RTRL complexity to the one of BPTT. However, this work did not consider RL.

Recently, \citet{irie2023} investigated a range of RNN architectures for which the RTRL updates can be computed efficiently. They assessed the practical performance of RTRL-based recurrent RL on a set of memory tasks, using modified LSTMs, and were able to show an improvement over training with BPTT when used in the framework of IMPALA \citep{espeholt2018}. Noteworthy, IMPALA requires collecting complete episodes before computing updates making it biologically implausible.

Finally, a great number of recent publications deal with training recurrent networks of spiking neurons (RSNNs). \citep{bellec2020, taherkhani2020, pan2023} The different approaches to train RSNNs in a biologically plausible manner do mostly rely on discrete spike events, for example in spike-time dependent plasticity (STDP). The e-prop algorithm introduced by \citet{bellec2020} stands out as the most similar to RFLO. It features the same computational complexity and has been shown to be capable of solving RL tasks, albeit only for discrete action spaces.

\section{Discussion}
We introduced real-time recurrent reinforcement learning (RTRRL), a novel approach to solving discrete and continuous control tasks for POMDPs, in a biologically plausible fashion. 
We compared RTRRL with PPO, which uses biologically implausible BPTT for gradient-computation. Our results show, that RTRRL with LRUs outperforms PPO consistently when using the same number of neurons. We further found, that using approximate gradients as in RFLO and FA, can still find satisfactory solutions in many cases. 

Although the results presented in this paper are empirically convincing, some limitations have to be discussed. The algorithm naturally suffers from a large variance when using a batch size of 1. It moreover needs careful hyperparameter tuning, especially when dealing with continuous actions.

RTRRL is grounded in neuroscience and can adequately explain how biological neural networks learn to act in unknown environments. The network structure resembles the interplay of dorsal and ventral striatum of the basal ganglia, featuring global RPEs found in dopaminergic pathways projecting from the ventral tegmental area and the substantia nigra zona compacta to the striatum and cortex~\citep{wang2018}. The role of dopamine as RPE was established experimentally by \citet{wise2004} who showed that dopamine is released upon receiving an unexpected reward, reinforcing the recent behavior. 
If an expected reward is absent, dopamine levels drop below baseline - a negative reinforcement signal. 
Dopaminergic synapses are usually located at the dendritic stems of glutamate synapses \citep{kandel2013} and can therefore effectively mediate synaptic plasticity. 
More specifically, the ventral striatum would correspond to the critic in RTRRL and the dorsal striatum to the actor, with dopamine axons targeting both the ventral and dorsal compartmens \citep{sutton2018}. The axonal tree of dopaminergic synapses is represented by the backward weights in RTRRL. 
Dopamine subsequently corresponds to the TD-error as RPE, which is used to update both the actor and the critic. RTRRL can therefore be seen as a model of reward-based learning taking place in the human brain.

Finally, an important reason for investigating online training algorithms of neural networks is the promise of energy-efficient neuromorphic hardware. \citep{zenke2021} The aim is to create integrated circuits that mimic biological neurons. Importantly, neuromorphic algorithms require biologically plausible update rules to enable efficient training. 

\section*{Acknowledgments}
Julian Lemmel is partially supported by the Doctoral College Resilient Embedded Systems (DC-RES) of TU Vienna and the Austrian Research Promotion Agency (FFG)
Project grant No. FO999899799. Computational results have been achieved in part using the Vienna Scientific Cluster (VSC).
\bibliography{aaai25}

\newpage
\clearpage
\appendix

\section{Real-Time Recurrent Reinforcement Learning Appendix}

\section{Derivation of update equations}
\label{sec-A-der}

Consider a CT-RNN that has $N$ hidden states and $I$ inputs, activation $\varphi$ and a combined weight matrix $W \in \R^{N\times X}$ where $Z\,{=}\,I\,{+}\,N\,{+}\,1$. Each neuron has a time-constant $\tau\,{\in}\,\R^{N}$ and the next state $h_{t+1}\in \R^{N}$ is computed as follows:
$$
h_{t+1} = h_{t}+\frac{1}{\tau}\left(-h_{t}+\varphi(W \xi_t)\right) \qquad \xi_t=\begin{bmatrix}  x_{t} \\h_{t}\\ 1\end{bmatrix} \in \R^{Z}
$$
where $x_{t}$ is the input at time $t$, and $1$ concatenated to $\xi_t$ accounts for the bias. The output $\hat y_{t} \in \R^O$ is given by a linear mapping $\hat y_{t}=W_{out}h_t$. The latent space follows the ODE $\tau\dot h = -h_{t}+\varphi(W \xi_t)$.

RFLO leverages the state-update expression in order to simplify the RTRL update. For this we expand the gradient of $f$ in equation \ref{eq:RTRL}. Note that previous work on RFLO kept the time constant $\tau$ fixed and trained the recurrent weights $W$ only, hence the restricted $\hat J^W$ in the equation:
\begin{equation} 
\label{eq:RFLO_der}
\begin{aligned}
\hat J_{t+1}^W =&\ \frac{d}{dW} h_{t} +\frac{d}{dW} f(x_{t},h_{t})\\
= &\ \frac{d}{dW} h_{t} + \frac{d}{dW} \frac{1}{\tau} \left(-h_{t}+\varphi(W \xi_t)\right)\\
= &\ \frac{d}{dW} h_{t} (1 -\frac{1}{\tau} ) + \frac{d}{dW} \frac{1}{\tau} \varphi(W \xi_t)\\
 = &\ (1 -\frac{1}{\tau})\ \hat J _t^W + \frac{1}{\tau} \nabla_W \varphi(W \xi_t)\\ 
 & \quad + \frac{1}{\tau} \hat J_{t}^W \nabla_{h_t} \varphi(W \xi_t)
\end{aligned}
\end{equation}
In order to achieve biological plausibility, RFLO boldly drops the last summand, since it requires horizontal gradient communication. The partial derivative of the activation is simply $\nabla_W \varphi(W \xi_t) = \varphi'(W \xi_t)^\top \xi_t$ where $\varphi'$ is the point-wise derivative of the activation function $\varphi$:
\begin{equation} 
\label{eq:A-RFLO}
\hat J_{t+1}^W \approx (1 -\frac{1}{\tau})\ \hat J _t^W  + \frac{1}{\tau} \varphi'(W \xi_t)^\top \xi_t
\end{equation}

We analogously derive the RFLO update for $\tau$.  Again we drop the communicated gradients $ \hat J_{t}^\tau \nabla_{h_t} \varphi(W\xi_t)$ and arrive at the expression for $\hat J^\tau$.

\begin{equation} 
\label{eq:A-RFLO_tau}
\begin{aligned}
\hat J_{t+1}^\tau = &\ \hat J_t^\tau - \frac{d}{d\tau} \frac{1}{\tau}h_{t}+\frac{1}{\tau}\varphi(W \xi_t)\\ 
= &\ \hat J_t^\tau - \nabla_\tau \frac{1}{\tau}h_{t} - \hat J_t^\tau  \nabla_{h_t}\frac{1}{\tau}h_{t} + \frac{d}{d\tau} \frac{1}{\tau}\varphi(W \xi_t)\\ 
 = &\ \hat J_t^\tau (1-\frac{1}{\tau}) + \frac{1}{\tau^2}h_{t} + \nabla_\tau \frac{1}{\tau}\varphi(W \xi_t)\\
  & \quad + \hat J_{t}^\tau \nabla_{h_t} \frac{1}{\tau} \varphi(W \xi_t) \\
= &\ \hat J _t^\tau (1 -\frac{1}{\tau}) + \frac{1}{\tau^2}h_{t} - \frac{1}{\tau^2}\varphi(W \xi_t)\\
  & \quad + \hat J_{t}^\tau \nabla_{h_t} \frac{1}{\tau} \varphi(W \xi_t)
\end{aligned}
\end{equation}

\section{Implementation Details}
\label{A-impl}
\begin{figure*}[!ht]
    \centering
    \hspace{-2em}
    \includegraphics[width=.6\linewidth]{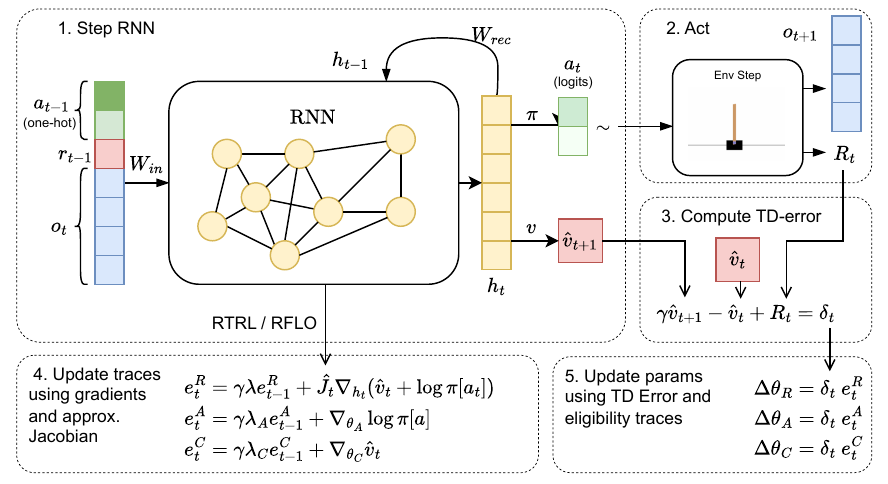}
    \includegraphics[width=.4\linewidth]{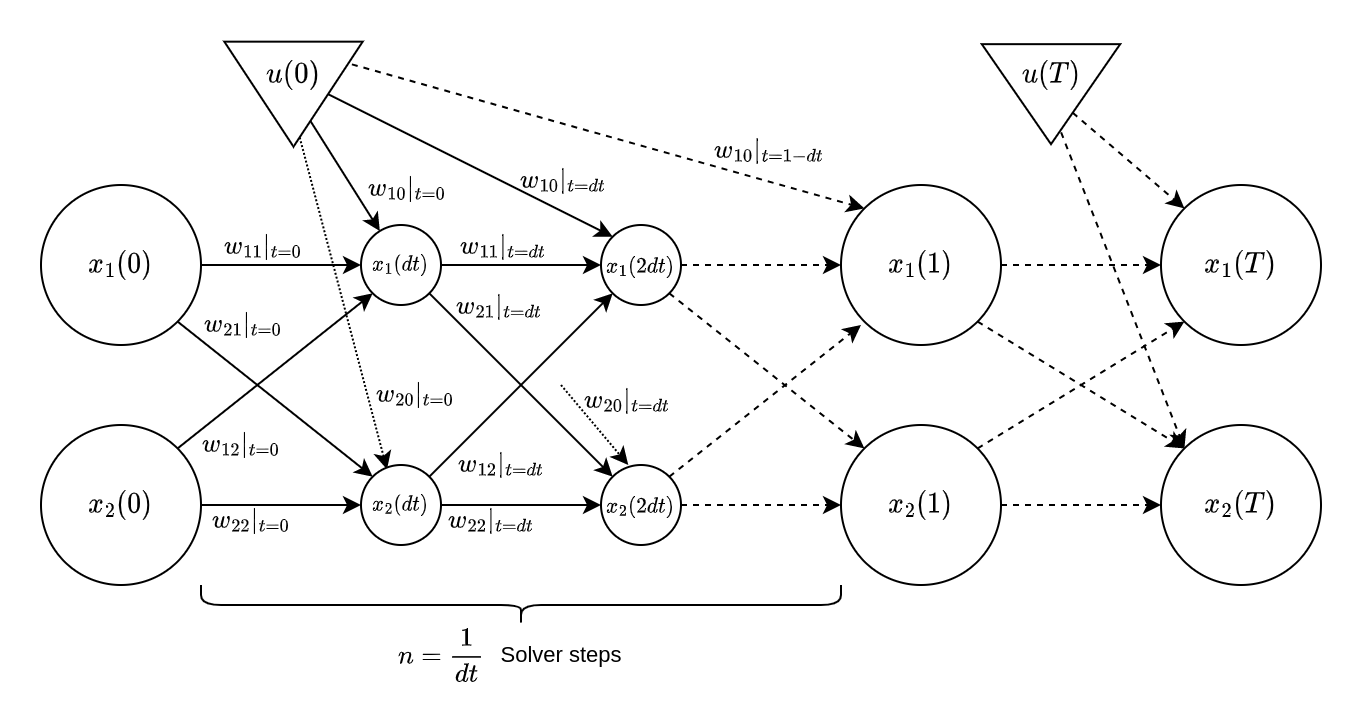}\hspace{-3em}
    \caption{Left: RTRRL can be devided into 4 parts that are repeated throughout training. Right: CT-RNNs are solving an Ordinary Differential Equation. In general, any solver may be used. When using forward Euler, $dt = k^{-1}$ is a hyperparameter that determines the number of solver steps and subsequently the accuracy of the solution.}
    \label{fig:steps}
\end{figure*}

Upon acceptance, we will publish our code on GitHub. Logging of experiments is implemented for Aim \footnote{https://aimstack.readthedocs.io/} or Weights \& Biases \footnote{httpbs://wandb.ai/} as backend. Our implementation is highly configurable and allows for many tweaks to the base algorithm. Available options include gradient clipping, learning rate decay, epsilon greedy policy, delayed RNN parameter updates and many more. Please refer to table \ref{tab:hparams} and the \texttt{Readme.md} in the code folder for a list of configurables. Table \ref{tab:hparams} summarizes the hyperparameters of RTRRL. We kept them at the listed default values for all our experiments.

\begin{table}
\begin{tabular}{l|c|c}
Description & Symbol & Value \\
\hline
number of neurons & $n$ & 32 \\
discount factor & $\gamma$ & 0.99 \\
Actor learning rate & $\alpha_A$ & 1e-4 \\
Critic learning rate & $\alpha_C$ & 1e-4 \\
RNN learning rate & $\alpha_R$ & 1e-4 \\
actor RNN trace scale & $\eta_A$ & 1.0 \\
entropy rate & $\eta_H$ & 1e-5 \\
Actor eligibility decay & $\lambda_A$ & 0.9 \\
Critic eligibility decay & $\lambda_C$ & 0.9 \\
RNN eligibility decay & $\lambda_R$ & 0.9 \\
CT-RNN ODE timestep & $dt$ & 1.0 \\
\multicolumn{2}{l|}{patience in epochs} & 20 \\
\multicolumn{2}{l|}{maximum environment steps} & 50 mil. \\
\multicolumn{2}{l|}{optimizer} & adam \\
\multicolumn{2}{l|}{batch size} & 1 \\
\multicolumn{2}{l|}{learning rate decay} & 0 \\
\multicolumn{2}{l|}{action epsilon} & 0 \\
\multicolumn{2}{l|}{update period} & 1 \\
\multicolumn{2}{l|}{gradient norm clip} & 1.0 \\
\multicolumn{2}{l|}{normalize observations} & False \\
\hline
\end{tabular}
\caption{Hyperparameters of RTRRL}
\label{tab:hparams}
\end{table}

\begin{figure*}[htb]
  \centering
  \includegraphics[width=.8\linewidth]{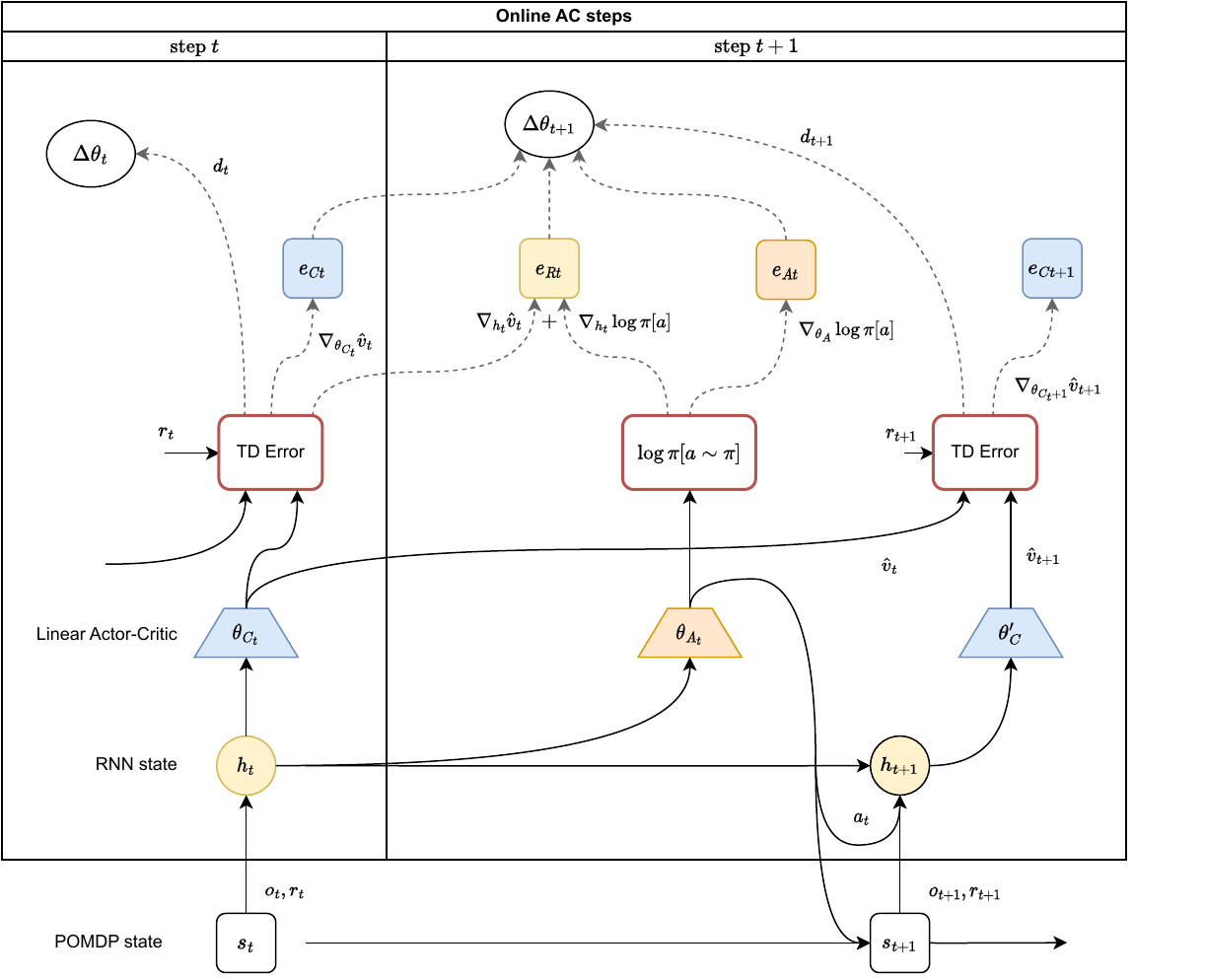}
  \caption{Flowchart depicting the computations done throughout one RTRRL step. }
  \label{fig:flow}
\end{figure*}

Algorithm \ref{alg:RTRRL_A} repeats RTRRL with if-cases for RTRL without Feedback Alignment for demonstrative purposes. The algorithm can be divided into 4 distinct steps that are depicted in figure \ref{fig:steps}. When developing the algorithm, we had to figure out the proper order of operations. Figure \ref{fig:flow} is a flowchart that was created to help understand at what point the eligibility traces are combined with the TD-error and approximate Jacobian, to form the parameter updates.

Since not using batched experiences, our algorithm unsurprisingly suffers from large variance and in some cases catastrophic forgetting ensues. Using an exponentially decaying learning rate for the RNN can help ins such cases but for simplicity we chose to make our experiments without this fix. Since a biologically plausible agent should retain its capability of reacting to shifts in the environment, the adaptability of our algorithm would be tainted as the learning rate of the RNN approaches 0. Nonetheless, RTRRL most of the time converges to an optimal solution without the use of decaying learning rates.

\paragraph{Neuron Model.}

The simplified CT-RNN outlined in the paper is taken from \citet{murray2019}. Our code allows for increasing the number of steps $k = dt^{-1}$ when solving the underlying Ordinary Differential Equation with the forward Euler method. More steps lead to a more expressive model meaning you can get away with fewer neurons, but also to increased computational complexity. In our experiments we kept $k=1$ for simplicity.

\begin{algorithm}[htbp]
    \caption{Real-Time Recurrent Reinforcement Learning with precise backpropagation and entropy regularization}
    \label{alg:RTRRL_A}
    \begin{algorithmic}[1]
	    \REQUIRE Linear policy: $\pi(a|h,\theta_{A})$
	    \REQUIRE Linear value-function: $\hat v(h,\theta_{C})$
	    \REQUIRE CT-RNN body: RNN$([o, a, r], h, \hat J, \theta_{R})$
	    \STATE $\theta_{A}, \theta_{C}, \theta_{R} \gets $ Randomly initialize parameters
	    \IF{RFLO}
		    \STATE $B_{A}, B_{C} \gets $ Randomly initialize feedback matrices
		\ENDIF
		\STATE $h, e_{A}, e_{C}, e_R \gets \mathbf 0$ 
		\STATE $o \gets $ Reset Environment
	    \STATE $h, \hat J \gets $RNN$([o, \mathbf 0, 0], h, \theta_{R})$
	    \STATE $v \gets \hat v(h, \theta_{C})$
		\WHILE{not done}
			\STATE logits $\gets \pi(h, \theta_{A})$
		    \STATE $a \gets $ Sample(logits)
		    \STATE $o, R \gets$ Take action $a$ in Environment
			\STATE $h', \hat J' \gets $RNN($[o, a, r], h, \hat J, \theta_{R}$) 
		    \STATE $v' \gets \hat v(h', \theta_{C})$
			\STATE $\delta \gets R + \gamma v' - v$
			\STATE $e_{C} \gets \gamma \lambda_{C} e_{C} + \nabla_{\theta_{C}} \hat v$
			\STATE $e_{A} \gets \gamma \lambda_{A} e_{A} + \nabla_{\theta_{A}} \ln \pi[a]$
			\IF{RTRL}
				\STATE $e_R \gets \gamma \lambda_{R} e_{R} + \nabla_{\theta_{R}} [\hat v + \eta_\pi \ln \pi[a]]$
                \STATE $g_{H,R} \gets \eta_{H} \nabla_{\theta_{R}} H(\pi)$
			\ELSIF{RFLO}
				\STATE $g_{C} \gets B_{C} \mathbf 1$
				\STATE $g_{A} \gets B_{A} \nabla_{h} \ln \pi[a]  $
				\STATE $e_R \gets \gamma \lambda_{R} e_{R} + \hat J  (g_{C} + \eta_A g_{A})$  
                \STATE $g_{H,R} \gets \eta_{H} \hat J \nabla_{h} H(\pi)$
                
			\ENDIF
			\STATE $\theta_{C} \gets \theta_{C} + \alpha_{C} \delta e_{C}$  
			\STATE $\theta_{A} \gets \theta_{A} + \alpha_{A} (\delta e_{A} + \eta_{H} \nabla_{\theta_{A}} H(\pi))$
			\STATE $\theta_{R} \gets \theta_{R} + \alpha_{R} (\delta e_R + g_{H,R})$
			\STATE $v \gets v', \quad h \gets h', \quad \hat J \gets \hat J'$
		\ENDWHILE
    \end{algorithmic}
\end{algorithm}

\section{More Experiments}

\begin{table*}[ht]
  \centering
  \small
\begin{tabular}{lcccc}
\toprule
Model & Linear TD($\lambda$) & RTRL CTRNN & RFLO CTRNN & PPO CTRNN \\
Environment &  &  &  &  \\
\midrule
CartPole-vel & 454.55$\pm$34.47 & 500.00$\pm$0.00 & 500.00$\pm$0.00 & 112.11$\pm$40.38 \\
CartPole-pos & 57.47$\pm$1.38 & 178.57$\pm$93.04 & 500.00$\pm$163.17 & 70.76$\pm$11.42 \\
MetaMaze & 28.00$\pm$43.33 & 34.00$\pm$3.28 & 40.00$\pm$7.30 & 10.80$\pm$1.34 \\
DiscountingChain & 1.10$\pm$0.04 & 1.05$\pm$0.04 & 1.10$\pm$0.05 & 1.10$\pm$0.00 \\
UmbrellaChain & 0.79$\pm$0.05 & 0.99$\pm$0.15 & 1.29$\pm$0.07 & 1.04$\pm$0.12 \\
BernoulliBandit & 404.40$\pm$20.51 & 373.80$\pm$18.20 & 371.70$\pm$5.17 & 89.14$\pm$18.54 \\
GaussianBandit & 0.01$\pm$5.83 & -126.17$\pm$70.60 & 24.05$\pm$21.43 & 0.00$\pm$0.00 \\
Reacher & 106.66$\pm$6.04 & 114.45$\pm$9.40 & 111.06$\pm$26.76 & 16.42$\pm$0.95 \\
Swimmer & 91.26$\pm$7.77 & 97.57$\pm$4.80 & 442.54$\pm$225.73 & 31.06$\pm$2.35 \\
MountCarCont & -30.00$\pm$8.94 & -40.00$\pm$5.48 & -0.03$\pm$0.67 & -1104.72$\pm$446.12 \\
MountCarCont-vel & -41.88$\pm$16.54 & -1.31$\pm$16.14 & -20.00$\pm$40.80 & -1079.70$\pm$204.52 \\
MountCarCont-pos & -46.91$\pm$15.50 & 0.04$\pm$8.43 & -59.54$\pm$43.45 & -1071.67$\pm$40.27 \\
MemoryChain-4 & 0.07$\pm$0.01 & 1.00$\pm$0.11 & 1.00$\pm$0.00 & 1.00$\pm$0.00 \\
MemoryChain-8 & 0.07$\pm$0.02 & 0.90$\pm$0.25 & 1.00$\pm$0.00 & 1.00$\pm$0.00 \\
MemoryChain-16 & 0.12$\pm$0.04 & 0.61$\pm$0.27 & 1.00$\pm$0.00 & 1.00$\pm$0.00 \\
MemoryChain-32 & 0.17$\pm$0.03 & 0.56$\pm$0.13 & 0.77$\pm$0.18 & 1.00$\pm$0.00 \\
MemoryChain-64 & 0.26$\pm$0.07 & 0.61$\pm$0.13 & 0.32$\pm$0.08 & 1.00$\pm$0.47 \\
DeepSea-4 & 0.99$\pm$0.00 & 0.99$\pm$0.44 & 0.99$\pm$0.00 & 0.99$\pm$0.00 \\
DeepSea-8 & 0.99$\pm$0.54 & 0.99$\pm$0.44 & 0.99$\pm$0.54 & 0.99$\pm$0.00 \\
DeepSea-16 & -0.00$\pm$0.54 & 0.99$\pm$0.54 & 0.99$\pm$0.44 & 0.00$\pm$0.00 \\
\bottomrule

\end{tabular}

\caption{Summary of RTRRL experiments on Gymnax for networks with 32 neurons. Numbers reported are the median and standard deviation of the best validation reward achieved throughout training for 5 runs (larger is better). "RTRRL RFLO" denotes our biologically plausible version of RTRRL.
  }
  \label{tab:results-gymnax}
  
\end{table*}

\begin{table*}[!ht]
  \centering
  \small

\begin{tabular}{lcccc}
\toprule
Model & Linear TD($\lambda$) & RTRRL RFLO & RTRRL LRU & PPO CTRNN \\
Environment &  &  &  &  \\
\midrule
MetaMaze & 55.93$\pm$43.33 & 61.60$\pm$9.24 & 15.80$\pm$5.02 & 38.79$\pm$27.03 \\
DiscountingChain & 1.09$\pm$0.02 & 1.10$\pm$0.00 & 1.10$\pm$0.00 & 1.32$\pm$0.00 \\
UmbrellaChain & 1.19$\pm$0.05 & 1.72$\pm$0.13 & 1.20$\pm$0.04 & 1.04$\pm$0.09 \\
Catch & 0.27$\pm$0.08 & 1.00$\pm$0.00 & 0.39$\pm$0.83 & 1.00$\pm$0.00 \\
BernoulliBandit & 393.38$\pm$20.51 & 893.96$\pm$11.35 & 807.96$\pm$100.19 & 883.44$\pm$18.72 \\
GaussianBandit & nan$\pm$nan & 61.10$\pm$32.98 & 37.78$\pm$67.01 & 0.00$\pm$0.00 \\
PointRobot & 1.34$\pm$0.21 & 5.30$\pm$4.30 & 0.98$\pm$0.49 & 1.04$\pm$0.57 \\
Reacher & 120.79$\pm$8.69 & 216.28$\pm$16.90 & 199.77$\pm$9.26 & 194.98$\pm$4.84 \\
Swimmer & 95.87$\pm$5.94 & 465.08$\pm$17.47 & 102.17$\pm$70.34 & 51.78$\pm$7.13 \\
MemoryChain & 0.08$\pm$0.01 & 1.00$\pm$0.00 & nan$\pm$nan & 1.00$\pm$0.00 \\
DeepSea & 0.99$\pm$0.00 & 0.99$\pm$0.00 & 0.99$\pm$0.00 & 0.99$\pm$0.00 \\
StatelessCartPole & 0.72$\pm$0.02 & 1.00$\pm$0.00 & 1.00$\pm$0.00 & 0.99$\pm$0.02 \\
NoisyStatelessCartPole & 0.50$\pm$0.01 & 0.99$\pm$0.01 & 0.93$\pm$0.04 & 0.59$\pm$0.03 \\
Minesweeper & 0.14$\pm$0.02 & 0.02$\pm$0.05 & 0.12$\pm$0.01 & -0.14$\pm$0.05 \\
MultiArmedBandit & 0.25$\pm$0.05 & 0.39$\pm$0.03 & 0.38$\pm$0.08 & 0.19$\pm$0.10 \\
RepeatFirst & 0.15$\pm$0.12 & -0.22$\pm$0.13 & 0.05$\pm$0.33 & 0.03$\pm$0.11 \\
RepeatPrevious & -0.35$\pm$0.01 & 0.54$\pm$0.23 & 1.00$\pm$0.00 & 0.93$\pm$0.01 \\
\bottomrule
\end{tabular}

\caption{Summary of RTRRL experiments including LRU experiments for networks with 32 neurons. Numbers reported are the median and standard deviation of the best validation reward achieved throughout training for 5 runs (larger is better). "RTRRL RFLO" denotes our biologically plausible version of RTRRL.
  }
  \label{tab:results}
  
\end{table*}

In Table \ref{tab:results} we summarize all our results for the experiments explained in the remainder of this section. The column "RFLO CT-RNN" corresponds to biologically plausible RTRRL as defined above, but we have also included results for RTRRL where RTRL is used in place of RFLO. The values presented in the table are the median and standard-deviation of the best evaluation results, throughout the training of 5 runs per environment and algorithm. Here, evaluation refers to running the environment for 10000 steps, which for example corresponds to 10 episodes for environments with a task duration of 1000 steps. As one can see, RTRRL performs best on average, followed by RTRRL with RTRL, as the second best algorithm.

In the remainder of this section we discuss some additional experiments we conducted that did not make it into the main manuscript of the paper.

\paragraph{Deep Exploration.}
Exploration versus exploitation is a trade-off, central to all agents learning a task online. The \texttt{DeepSea} environment included in \texttt{bsuite} \citep{osband2020} is tailored to benchmark the ability of RL algorithms to explore in unfavourable environments. 
The agent is required to explore to reach the goal position albeit receiving negative rewards when moving towards it. We give results for exponentially increasing task length.

\begin{figure}[ht]
  \centering
\includegraphics[width=.8\linewidth]{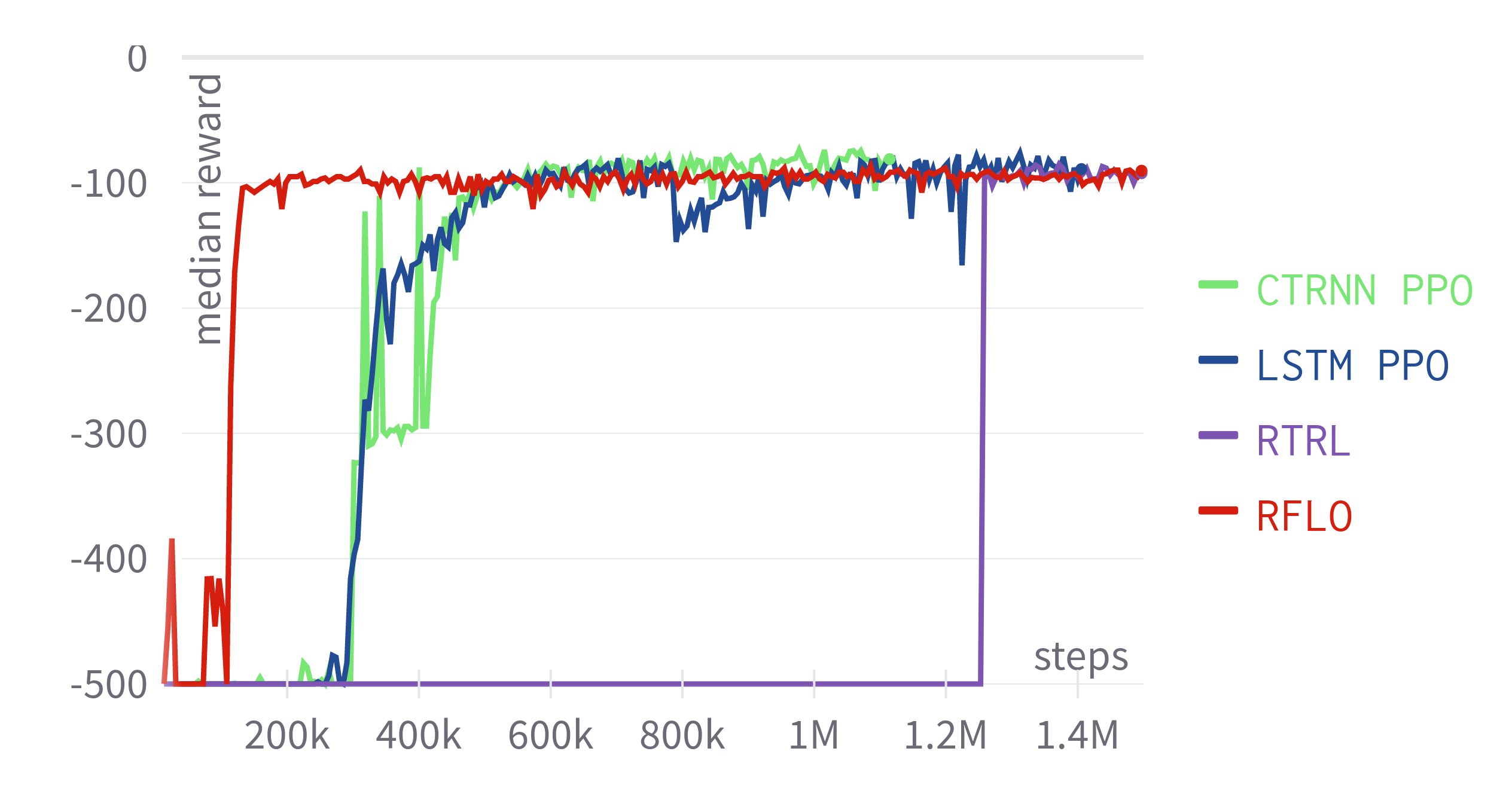}
\caption{Median reward of 5 runs on \texttt{Acrobot}. The key to solving this environment is exploration. RNNs trained with RFLO solve it quicker than when trained with BPTT.}
  \label{fig:acrobot}
\end{figure}

The classic control environment \texttt{Acrobot}, too, requires extensive exploration. A double pendulum starting in hanging position is set into motion by controlling the middle joint. The outer segment has to be elevated above a certain height, located above the anchor of the inner segment. Figure \ref{fig:acrobot} shows the median rewards of 5 runs over the number of environment steps. We find that RTRRL with RFLO finds a solution significantly faster than the other methods hinting at superior exploration of our RTRRL approach. 

\begin{figure*}
  \centering
\hspace{-0.5em}
\includegraphics[width=.5\linewidth]{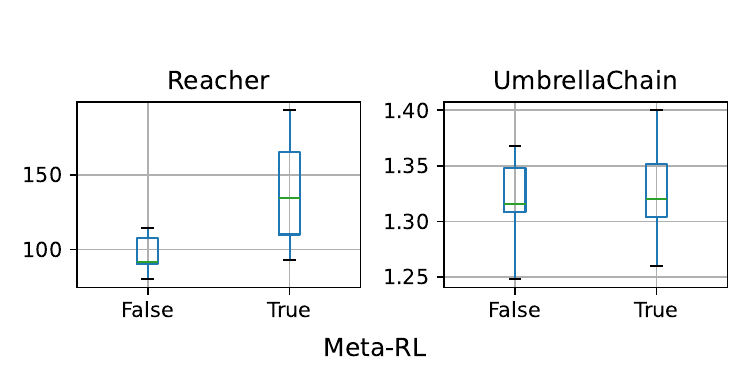}
\includegraphics[width=.5\linewidth]{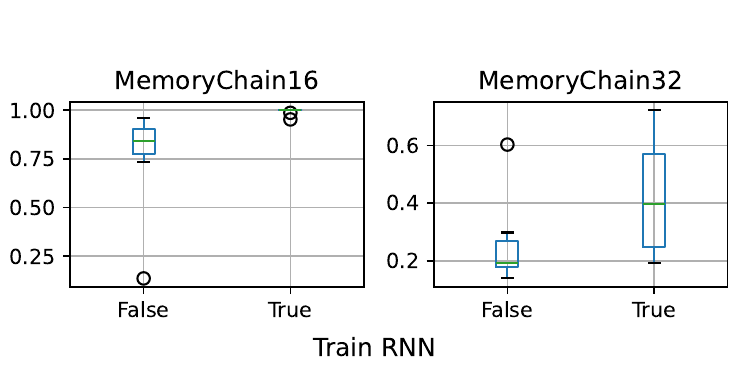}\hspace{-0.5em}

\includegraphics[width=1\linewidth]{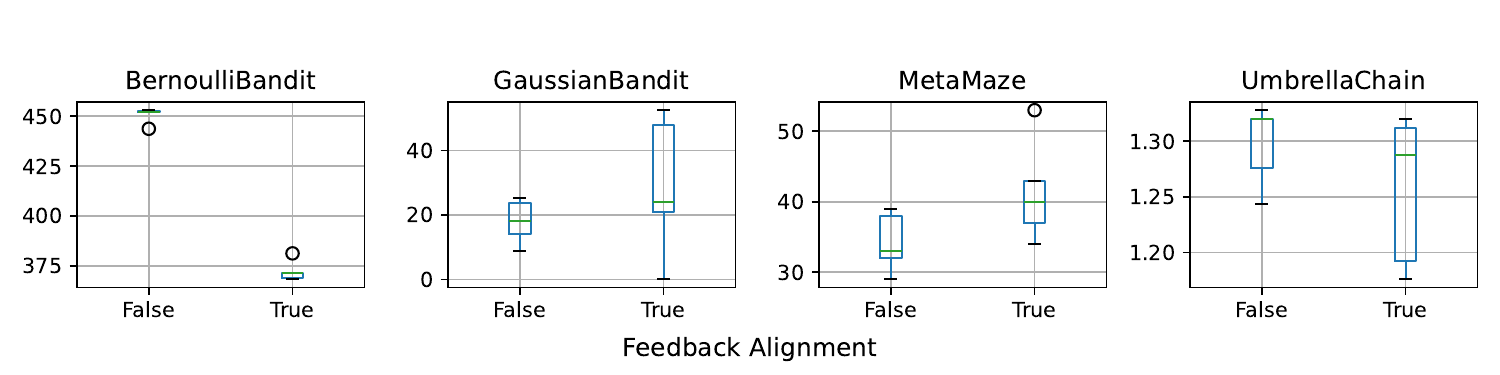}
  \caption{Left: While in most cases it does not make a difference, not using the Meta-RL architecture hampers performance in some cases. Right: Keeping the RNN frozen as in echo state networks leads to significantly worse results. Bottom: Comparing RTRRL with and without biologically plausible propagation of gradients using FA. Some environments such as \texttt{GaussianBandit} and \texttt{MetaMaze}, seem to benefit from it, while others do better with regular backpropagation.}
  \label{fig:multi}
\end{figure*}

\paragraph{Reservoir Computing.} The term reservoir computing refers to using a fixed RNN with trainable linear readout weights, sparing the hassle of exploding or vanishing gradients. \citet{yamashita2022} employed this technique for solving POMDPs. We conducted experiments to test whether optimizing the RNN in RTRRL is actually beneficial. By keeping the RNN fixed after initialization, training the linear actor and critic only, we create our own echo state network. Figure \ref{fig:multi} right shows results for the \texttt{MemoryChain} environment. We conclude that training the RNN does improve performance.

\section{Biological implausibility of BPTT}
\label{sec-A-bptt}

The three main objections for the plausibility of BPTT in biological neural networks are: First, the reliance on shared weights between the forward and the backward synapses; Second, reciprocal error-transport, which means propagating back the errors, without interfering with the neural activity \citep{bartunov2018}; Third, the concern that BPTT requires storing long sequences of the exact activation for each cell \citep{lillicrap2019a}. We rephrase the premises of \citet{bartunov2018} and \citet{lillicrap2019a} as follows. A biologically plausible learning algorithm has to be: 
(1) \textit{Local}, up to some low-dimensional reward signal, (2) \textit{Online}, computing parameter updates and the network outputs in parallel, and (3) \textit{Without weight transport}, in its error computation, meaning that synapses propagating back error signals must not mirror the strength of forward synapses. We further elaborate on each of these requirements for obtaining biological plausibility in the following.

{\it Fully online.} 
The computation of updates should not depend on alternating forward and  backward phases. Conceivably, such phases could be implemented using pacemaker neurons. However, there must not be any freezing of values, which occurs in BPTT. Furthermore, parallel streams of experience, such as in batched environments of modern DRLs, violate this constraint, as a biological agent can only interact with the singular environment in which it is situated.

{\it No weight transport.}
Synapses that propagate back error signals cannot have their strength mirroring the strength of forward synapses. This is heavily violated by backpropagation since backward pathways need access to, and utilize, parameters that were used in the forward pass.

{\it No horizontal gradient communication.}
Individual neurons $x_{k}$ are very likely unable to communicate exact activation gradients $\nabla_{w_{ij}}x_{k}(t)$, with respect to the synaptic parameters $w_{ij}$, to other neurons that are not in their vicinity. In other words, intermediate (source) neurons should not be able to tell other (target) neurons how each of the synaptic weights of the source influence the dynamics of the target, nor vice-versa.

\section{Biological Interpretation of RTRRL}

Real-Time Recurrent Reinforcement Learning is a model of goal-directed behaviour learning and training of reflexes in animals, located in the human basal ganglia \citep{rusu2020}. It comprises a scalar-valued global reward signal that can be interpreted as dopaminergic synapses, a value estimator (critic) and motor output (actor).

\subsection{Dopamine and Learning}

Dopamine is a neurotransmitter found in the central nervous system of mammals. It is widely agreed upon that dopamine plays an important role in rewards and reinforcement. The molecule is released by specialized neurons located primarily in two areas of the brain: the substantia nigra zona compacta (SNc) and the ventral tegmental area (VTA). \citep{wise2004} These neurons have large branching axonal arbors making synapses with many other neurons - mostly in the striatum and the pre-frontal cortex. Those synapses are usually located at the dendritic stems of glutamate synapses and can therefore effectively influence synaptic plasticity. \citep{sutton2018}

Whenever receiving an unexpected reward, dopamine is released which subsequently reinforces the behavior that led to the reward. However, if the reward is preceded by a conditioned stimulus, it is released as soon as the stimulus occurs instead of when receiving the expected reward. If then the expected reward is absent, levels will drop below baseline representing a negative reinforcement signal. \citep{wise2004} Various experiments have shown that dopamine is crucial for stamping in response-reward and stimulus-reward associations which in turn is needed for motivation when confronted with the same task in the future. Particularly, moderate doses of dopamine antagonists (neuroleptics) given to a live animal will reduce motivation to act. Habitual responses decline progressively in animals that are treated with neuroleptics. \citep{wise2004}

\subsection{Ternary Synapses}

Hetero-synaptic plasticity is a type of synaptic plasticity that involves at least three different neurons. Usually, a sensory neuron is forming a synapse with a motor neuron. A third so called \textit{facilitating} neuron also forms a synapse at the same spot and so is able to influence the signal transmission between the sensory and the motor neuron. A historic example is found in the gill-withdrawal reflex circuitry of Aplysia \citep{kandel2013}. The RPE signalling, facilitating inter-neuron strengthens the motor response (withdrawing the gill) when a negative reward (shock) is experienced or expected. RTRRL could be implemented in biological neural networks with ternary synapses where the facilitating synapses are projecting back from the TD-error computing inter-neuron.

\begin{figure}[ht]
\label{fig:basal-ganglia}
\centering
  \includegraphics[width=.8\linewidth]{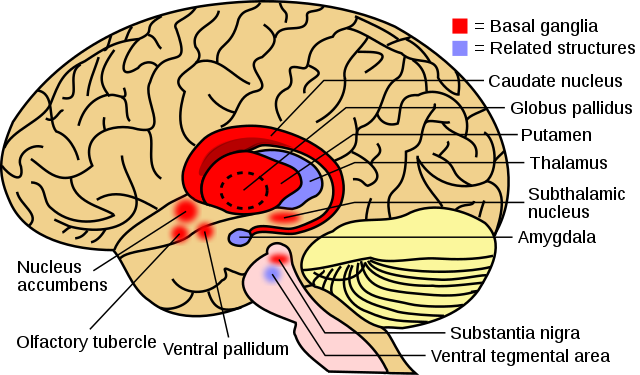}
  \caption[The basal ganglia are located at the base of the forebrain and play a major role in motivation and behavioural learning. Source: Wikimedia, Leevanjackson]{The basal ganglia are located at the base of the forebrain and play a major role in motivation and behavioural learning. Source: Wikimedia, Leevanjackson\footnotemark}
  
\end{figure}
\footnotetext{https://commons.wikimedia.org/w/index.php?curid=85845448}

\subsection{Brain Structures}

There is ample evidence that certain brain structures are implementing \emph{actor-critic} methods. The striatum is involved in motor and action planning, decision-making, motivation, reinforcement and reward perception. It is also heavily innervated by dopamine axons coming from the VTA and the SNc. It is speculated that dopamine release in the ventral striatum "energizes the next response" while it acts by stamping in the procedural memory trace in the dorsal striatum "establishing and maintaining procedural habit structures". \citep{wise2004} Subsequently, the ventral striatum would correspond to the \emph{critic} and the dorsal striatum to the \emph{actor} of an RL algorithm \citep{sutton2018}. Dopamine would then correspond to the \emph{TD-error} which is used to update both the \emph{actor} and the \emph{critic}. As dopamine neuron axons target both the ventral and dorsal striatum, and dopamine appears to be critical for synaptic plasticity, the similarities are evident. Furthermore, the \emph{TD-error} and dopamine levels are both encoding the RPE: they are high whenever an unexpected reward is received and they are low (or negative in case of the \emph{TD-error}) when an expected reward does not occur. \citep{sutton2018} These similarities could be beneficial for RL as well as for Neuroscience as advances in either field could lead to new insights that are beneficial to the other.

\section{Timing comparison}

In order to compare wall-clock time, we did a quick performance test for RTRRL with CT-RNN and PPO with LSTM. In both cases we trained for 5000 steps, with 32 units and a batch size of 1, on a machine with a single GeForce RTX 2070 GPU. For both cases, we repeated the test 3 times and calculated the average time per step of the algorithm. Our results were 7,58 ms / step for PPO-LSTM and 7,49 ms / step for RTRRL-RFLO. Please note that these results may not carry over to larger model or batch sizes.

\end{document}